\documentclass[sigconf, noacm]{acmart}

\newcommand{\yr}{\textcolor{black}}
\newcommand{\yrn}{\textcolor{violet}}
\usepackage{gensymb}
\begin{document}

\pagestyle{plain}
\renewcommand{\shortauthors}{}
\setcopyright{none}

\title{HeadLighter: Disentangling Illumination in Generative 3D Gaussian Heads via Lightstage Captures}

\author{
Yating Wang$^{1}$\textdagger \quad
Yuan Sun$^{2}$\textdagger \quad
Xuan Wang$^2$ \quad 
Ran Yi$^{1*}$ \quad
Boyao Zhou$^2$ \quad
Yipengjing Sun$^2$ \and 
Hongyu Liu$^2$ \quad
Yinuo Wang$^2$ \quad 
Lizhuang Ma$^{1*}$
\\
$^1$Shanghai Jiao Tong University 
\quad
$^2$AntGroup Research \\
}

\begin{abstract}
Recent 3D-aware head generative models based on 3D Gaussian Splatting achieve real-time, photorealistic and view-consistent head synthesis.
However, a fundamental limitation persists: the deep entanglement of illumination and intrinsic appearance prevents controllable relighting.
Existing disentanglement methods rely on strong assumptions to enable weakly supervised learning, which restricts their capacity for complex illumination.
To address this challenge, we introduce HeadLighter, a novel supervised framework that learns a physically plausible decomposition of appearance and illumination in head generative models. 
Specifically, we design a dual-branch architecture that separately models lighting-invariant head attributes and physically grounded rendering components. 
A progressive disentanglement training is employed to gradually inject head appearance priors into the generative architecture, supervised by multi-view images captured under controlled light conditions with a light stage setup.
We further introduce a distillation strategy to generate high-quality normals for realistic rendering.
Experiments demonstrate that our method preserves high-quality generation and real-time rendering, while simultaneously supporting explicit lighting and viewpoint editing.
We will publicly release our code and dataset.
\end{abstract}


\begin{teaserfigure}
  \includegraphics[width=\textwidth]{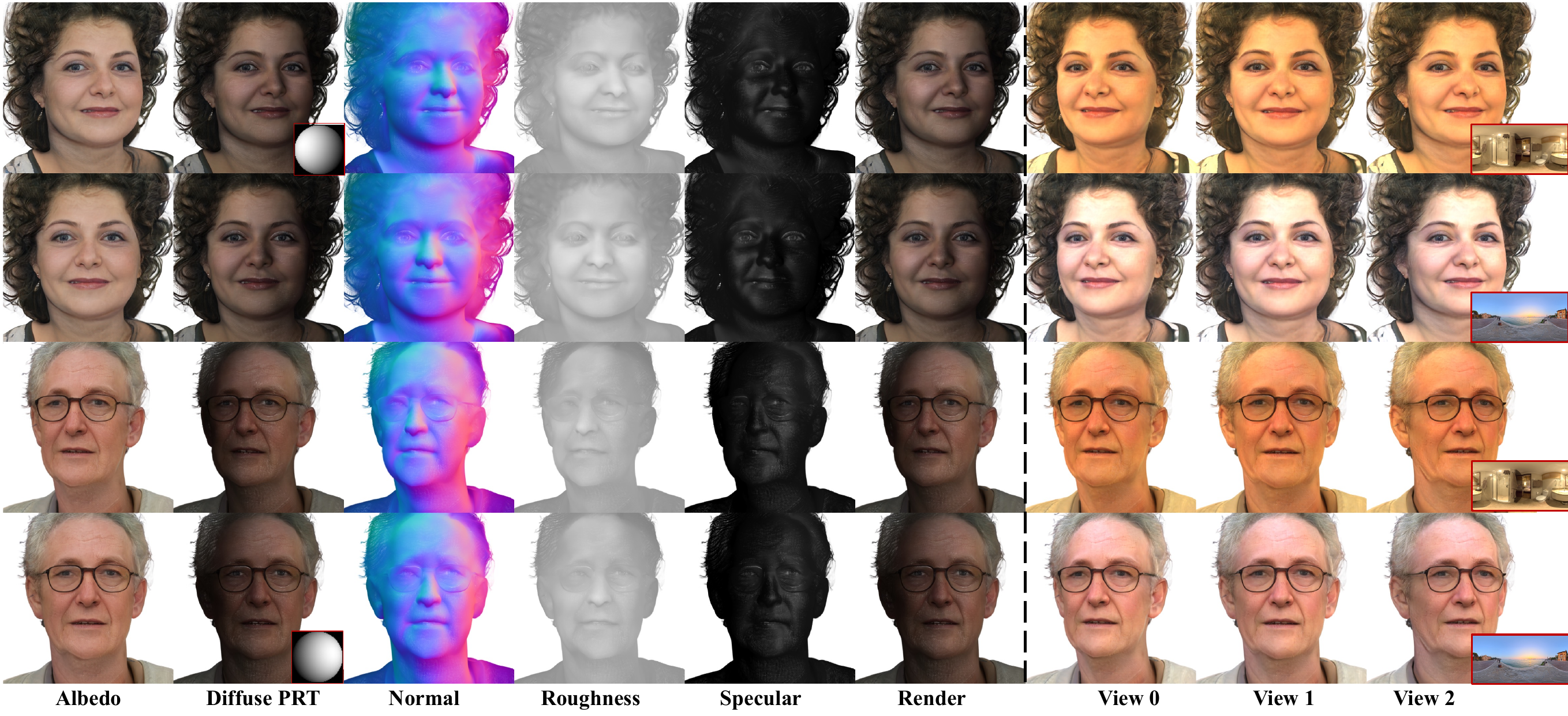}
  \vspace{-2.5em}
  \caption{Our method achieves high-fidelity disentanglement of intrinsic appearance and illumination in generative 3D Gaussian head models. It produces photorealistic intrinsic attributes including albedo/normals/roughness and utilizes precomputed radiance transfer (PRT) for diffuse global shading, and supports relighting under diverse lighting conditions such as environment maps, spherical harmonics. Left: intrinsic appearance attributes of two generated subjects (rendered under the SH lighting shown in the red boxes). Right: relighting results under varying environment maps(shown in the red boxes). }
  \label{fig:teaser}
\end{teaserfigure}

\maketitle

\section{Introduction}
Generative models for 3D human heads have seen remarkable progress, driven by their wide applications in digital avatars and virtual reality. Among them, the most prominent approaches are 3D-aware Generative Adversarial Networks (GANs) built upon neural rendering techniques 
~\cite{kerbl20233d, mildenhall2021nerf}. These models learn a rich prior of 3D head geometry and appearance directly from large-scale 2D in-the-wild images, without requiring any explicit 3D supervision. Consequently, they can synthesize highly realistic and view-consistent 3D head with rich identity and appearance diversity.
In the pursuit of real-time performance and impressive visual quality, recent works integrate 3D Gaussian Splatting (3DGS) into the generative framework~\cite{gghead, barthel2025cgs}
, replacing the computationally intensive volume rendering and achieving real-time rendering. 

However, a critical limitation persists: the illumination and the intrinsic head appearance (i.e., material or albedo) are deeply entangled, fundamentally preventing fine-grained control over the lighting conditions of the generated 3D heads.
Existing approaches enable unsupervised decomposition of pretrained generative models using simplified reflectance models or strong assumptions. 
NFL ~\cite{nerffacelighting} involves style-mixing supervision with the white-light and face symmetry assumption, initializing from an on-the-shell light estimator, while GSHR~\cite{lv2025gsheadrelight} introduces a simplified light transport function for 3D gaussian and distills albedo and shading from NFL~\cite{nerffacelighting}.
Others~\cite{deng2024lumigan, ranjan2023facelit} enforce physically plausible shading in neural volume rendering to split the diffuse and specular components. 
While these methods support explicit relighting, lack of real-world data supervision limits their ability to handle complex lighting and achieve highly physical plausibility.

We address this by an intuitive and straightforward way: involving a multi-view face dataset captured under controlled illuminations with a LightStage to learn the disentanglement.
However, directly utilizing such dataset is non-trivial: compared to large-scale in-the-wild face image datasets, our light stage dataset covers only a limited range of identities and lighting conditions, and exhibits a noticeable domain gap relative to natural imagery.
To prevent overfitting on the lightstage dataset and preserve the generalization capability of the base generative model, we decompose pretrained generativte model into a dual-branch architecture: \textbf{1)} a light-invariant base branch to generate geometry and albedo of 3D gaussian, and \textbf{2)} a relight branch predicting attributes for physically-based rendering.
To ensure stable disentanglement, we devise a progressive training strategy. 
First, we obtain base branch by fine-tuning the intermediate layers of a pretrained generator to remove illumination while preserving generation diversity. 
Next, with the base branch frozen, we inject head material priors from LightStage captures into the relight branch. 
Finally, we enhance the model's generalization across diverse illumination environments by employing adversarial training with randomly sampled lighting conditions.

A further challenge lies in obtaining accurate surface normals, which are crucial for realistic shading but not native to 3D gaussian representations. 
While computing normals from extracted mesh or neighbor splats are feasible, 
such methods are computationally expensive in training iterations and can introduce significant noise. 
Observing that the base branch encodes face geometry, we task the relight branch to directly predict normals guided by base branch features.
We further introduce a distillation strategy and smoothness terms that enforce consistent and plausible normal prediction, enabling photorealistic rendering of complex shading effects. 

\begin{figure*}
    \centering
    \includegraphics[width=\textwidth]{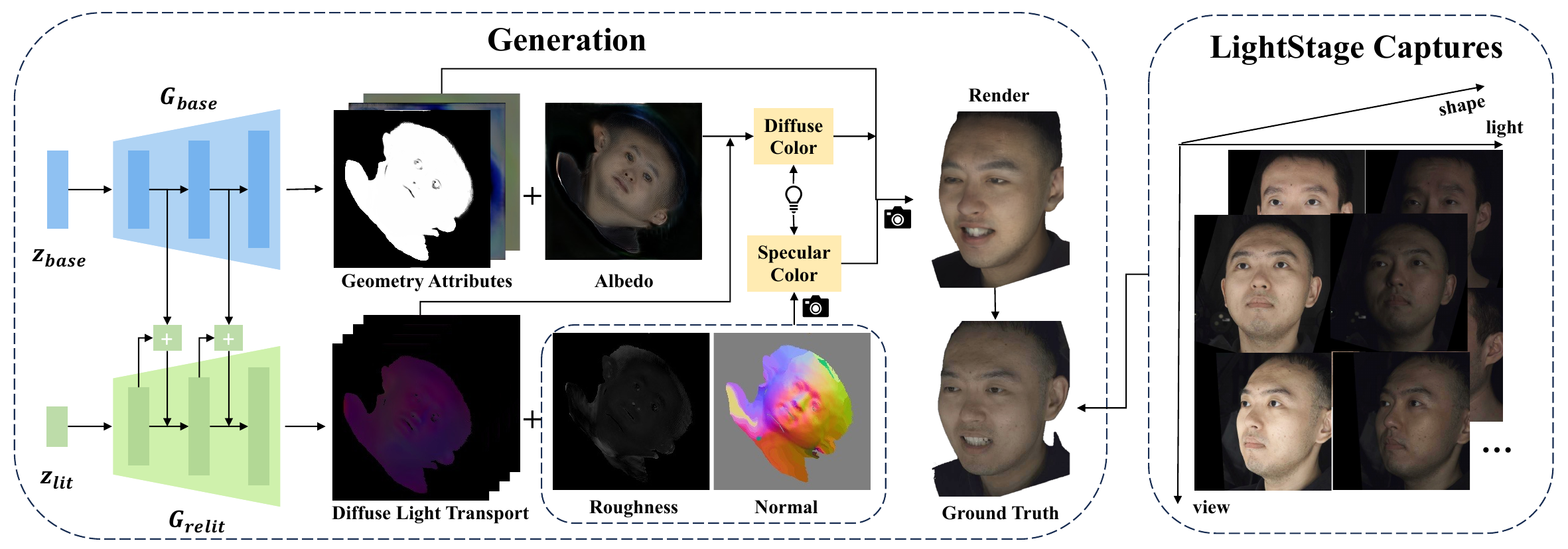}
    \vspace{-15pt}
    \caption{\textbf{Our proposed framework.} 
    We leverage real-world captures to learn high-quality disentanglement of illumination and appearance in generative 3D gaussian heads. 
    We capture multi-view images of different subjects under various light conditions via lightstage.
    A dual-branch architecture is employed to separately generate light-invariant gaussian attributes (geometry $\mathcal{A}_{geo}$ and albedo $\rho$), and physically-based rendering attributes (normal $\mathbf{n}_k$, roughness$\sigma_k$ and diffuse light transport coefficients $T_k$). 
    These attributes together with camera pose and light condition can be splatted into images and compared to the real-world captures to optimize the network.
    }
    \vspace{-10pt}
    \label{fig:pipeline}
\end{figure*}

We demonstrate the effectiveness of our pipeline through comprehensive comparisons with state-of-the-art methods. 
We evaluate generative diversity, rendering speed and multi-view spatial consistency of our disentangled generative models.
We demonstrate the rendering results of 3D human heads generated by our method under different lighting inputs.
To further validate the disentanglement capability, we quantitatively and qualitatively compare free-view relighting results by extracting image pairs from our real-captured datasets.
To foster future research, we will make our code and licensed dataset publicly available.
\section{Related Works}
\label{sec:rela}
\subsection{3D-aware Generative Head Models} The pursuit of high-fidelity, view-consistent, and controllable 3D human head synthesis has led to the rapid evolution of 3D-aware Generative Adversarial Networks (GANs). These models learn to generate 3D heads from large-scale 2D image sets, bypassing the need for explicit 3D supervision. Early 3D GANs integrate various 3D representations~\cite{eg3d, chan2021pi, schwarz2020graf, next3d, tang20233dfaceshop} into the GAN framework. Among them, EG3D~\cite{eg3d} utilizes a hybrid tri-plane representation, achieving photorealistic visual quality. However, its reliance on volume rendering is computationally expensive, and the super-resolution module can introduce view inconsistencies. To facilitate efficient, high-resolution generative head models, recent methods like \cite{barthel2025cgs, yu25gaia, gghead} have turned to 3D Gaussian Splatting (3DGS)~\cite{kerbl20233d}. However, a fundamental limitation persists: intrinsic face appearance properties and lighting are conflated into a view-dependent color, which renders direct and physically plausible relighting impossible.

To enable explicit lighting control, a line of research~\cite{pan2021shading, tan2022volux, nerffacelighting, deng2024lumigan, lv2025gsheadrelight, jiang2025nerffaceshop,ranjan2023facelit} has focused on disentangling illumination. NFL~\cite{nerffacelighting} distills pseudo albedo and shading from a pretrained EG3D~\cite{eg3d} model via style-mixing, constrained by white-light and face symmetry assumptions, along with an on-the-shelf light estimator. LumiGAN~\cite{deng2024lumigan} and FaceLit~\cite{ranjan2023facelit} employ an inverse rendering framework to separately model diffuse and specular components. GSHR~\cite{lv2025gsheadrelight} achieves highly efficient synthesis by designing a simplified radiance transfer representation for 3D Gaussians, but it relies on supervision from pseudo-data generated by NFL~\cite{nerffacelighting}. Consequently, these approaches lack access to ground-truth data and must resort to strong, simplifying assumptions that limit their ability to model the complex interaction between realistic materials and colored illumination. To address this, we propose HeadLighter, a high-fidelity relightable generative 3D Gaussian head model that learns illumination disentanglement from LightStage captures.

\subsection{Portrait Relighting} Portrait relighting~\cite{yeh2022learning, hou2022face, pandey2021total, cai2024real, guo2025high, mei2023lightpainter, feng2021learning} aims to render a person's face under novel lighting conditions. 
The core challenge lies in solving an ill-posed inverse rendering problem to separate intrinsic properties from the scene's illumination. 
Existing works~\cite{debevec2000acquiring, rgca} capture One-Light-at-a-Time (OLAT) images with LightStage to enable rendering under novel lighting, yielding high-fidelity results. However, they require controlled capture settings and do not generalize to in-the-wild, sparse-view portraits. 
To enable relighting of in-the-wild portraits, subsequent methods introduce various priors and assumptions. Early works~\cite{zhou2019deep, ponglertnapakorn2023difareli, split2023} rely on low-rank priors like 3D Morphable Models~\cite{blanz2023morphable}. Others~\cite{wang2020single, chaturvedi2025synthlight, yeh2022learning} leverage synthetic datasets for training but face a sim-to-real domain gap. SwitchLight~\cite{kim2024switchlight} learns diffusion prior from OLAT data and synthetic data to enable single-image portrait relighting.
Similarly, URAvatar~\cite{li2024uravatar} learns shape and appearance priors from large-scale OLAT data to support relightable reconstruction from smartphone videos.
Some works~\cite{cai2024real, Guo_2025_CVPR, Mei_2025_CVPR} address portrait video relighting, focusing on consistency across frames. Moreover, some methods also attempt relighting for both human body~\cite{Wang_2025_CVPR, Teufel_2025_ICCV} and general scenes~\cite{zhang2025scaling, Xing_2025_CVPR, bharadwaj2025genlit}.

Another line of research has shifted towards leveraging powerful generative priors. Initial attempts treated relighting as a style transfer task within the latent space of 2D GANs~\cite{abdal2021styleflow, deng2020disentangled, yi2024feditnet++, br2021photoapp, shoshan2021gan}, but struggled to decouple illumination from other facial attributes like pose and expression. 
Recently, a growing body of work~\cite{lite2relight,holorelight,deng2024lumigan,pan2021shading,tan2022volux,nerffacelighting,lv2025gsheadrelight, rao20253dpr} imposes generative 3D face priors to jointly support relighting and novel-view synthesis. HoloRelight~\cite{holorelight} utilizes OLAT and synthetic data for training, first delighting the input and then applying new lighting conditions via an implicit shading representation. 
3DPR~\cite{rao20253dpr} enables generating a basis of virtual OLAT images from a single portrait and then recombining them for relighting. 
Similar to NFL~\cite{nerffacelighting} and GSHR~\cite{lv2025gsheadrelight}, our method focuses on relightable 3D head generation but can also perform portrait relighting via an inversion-then-relight pipeline; however, under OLAT captures supervision, it achieves more physically accurate lighting disentanglement and thus better relighting quality.
\section{Methods}
\subsection{Preliminaries}
\noindent\textbf{3D Head GAN}
leverages neural rendering techniques to learn priors of 3D facial shape and appearance from large-scale 2D image collections. To overcome the computational bottleneck of traditional volumetric rendering, recent works integrate the 3D GAN framework with fast rasterization of 3D Gaussian Splatting (3DGS)~\cite{kerbl20233d}. 
A prominent example is GGHead~\cite{gghead}, which employs a StyleGAN2-based generator~\cite{karras2019style} to map a latent code $z \in \mathbb{R}^{512}$ to multi-channel 2D attribute maps $M_{\star}$ defined in the UV space of a template head mesh. 
The attributes for 3D Gaussian $\{G_k\}$, can be obtained by sampling the feature maps $M_{\star}$ at UV coordinate $x_k^{uv}$.
These attributes are categorized into: (1) geometry properties, including position $\boldsymbol{\mu}_k \in \mathbb{R}^3$, rotation $\mathbf{q}_k \in \mathbb{R}^4$, scale $\mathbf{s}_k \in \mathbb{R}^3$, and opacity $o_k \in \mathbb{R}$; and (2) view-dependent radiance defined with 1-degree Spherical Harmonics (SH) $a_k \in \mathbb{R}^{3\times(1+1)^2}$. The framework is further stabilized with several regularization terms during adversarial training.

\subsection{Data Acquisition}
We capture multi-view One-Light-A-Time(OLAT) images using LightStage system equipped with 46 calibrated uniform distributed point lights and 28 synchronized cameras. Both the lights and cameras operate at 25 fps.
We record under various lighting configurations, including a “\textbf{Uniform}” mode with all 46 lights \yr{turning} on, a “\textbf{Direction}” mode where each light is activated together with its neighboring lights, and “\textbf{Random}” mode where 10 or 20 lights are randomly selected. 
To acquire accurate 2D detection for each frames, we insert a fully illuminated frame every 10 frames, assume the subject remains stationary across these 11 frames, and propagate its estimation to the intervening frames.
We capture three expressions for each subject: neutral, eyes closed, and smile.
More details about the dataset processing can be found in \ref{sec:dataset}.

\subsection{Relightable 3DGS Head Generation}
We represent a relightable 3D head as a collection of 3D Gaussians $\{g_k\}$, where each Gaussian $g_k$ is associated with a set of geometric and appearance attributes and can be rendered into an image via splatting in a differentiable, physically-based manner, given arbitrary lighting conditions and \yr{viewpoints}.
All attributes of the 3D Gaussians are embedded into structured 2D UV maps of a template head mesh, from which each Gaussian \( g_k \) samples its attributes at its assigned UV coordinate. 
We generate the geometry and material maps using a dual-branch StyleGAN generator. The two branches share a common UV parameterization but take distinct latent codes to disentangle geometry from lighting-specific attributes:
\begin{itemize}
    \item \textbf{Base branch} $G_{\text{base}}: \mathbf{z}_{\text{base}} \to (\mathcal{A}_{\text{geo}}, \boldsymbol{\rho})$. It maps a latent code $\mathbf{z}_{\text{base}} \in \mathbb{R}^{512}$ to the geometric attributes $\mathcal{A}_{\text{geo}} = \{ \boldsymbol{\mu}_k, \mathbf{s}_k, \mathbf{q}_k, o_k \}$ and albedo $\boldsymbol{\rho}_k$.
    
    \item \textbf{Relight branch} $G_{\text{relit}}: \mathbf{z}_{\text{lit}} \to (\mathbf{n}, \sigma, \mathbf{T})$. It takes a separate latent code $\mathbf{z}_{\text{lit}} \in \mathbb{R}^{128}$ to predict relighting-related attributes, including surface normals $\mathbf{n}_k \in \mathbb{R}^3$, scalar roughness $\sigma_k$, and diffuse light transport coefficients $\mathbf{T}_{k} \in \mathbb{R}^{3 \times (s+1)^2}$, where $s$ denotes the SH degree.
\end{itemize}

Given incident lighting $\mathcal{L} = \{(\omega_i, \mathbf{I}_i)\}_{i=1}^M$, where $\omega_i$ denotes the direction of the $i$-th point light, and $\mathbf{I}_i \in \mathbb{R}^3$ \yr{denotes} RGB radiance (encoding both color and intensity), the final color $c_k$ of each Gaussian is computed as the sum of diffuse component $c_{\text{diffuse}, k}(\mathcal{L})$ and specular component $c_{\text{specular}, k}(\mathcal{L}, \omega_o)$, where $\omega_o$ is the viewing direction. 
Then \yr{the final color} $c_k$ \yr{and} geometry attributes $\mathcal{A}_{\text{geo}}$ \yr{of 3D Gaussians} can be rendered into images via efficient splatting, enabling high-fidelity, and relightable head synthesis under arbitrary illumination.

\yr{Specifically,} for the view-independent \textbf{diffuse} term \yr{$c_{\text{diffuse}, k}$}, we adopt Precomputed Radiance Transfer (PRT)~\cite{sloan2002precomputed} to model global illumination effects, including self-occlusion and subsurface scattering. The illumination from each light source is projected into SH coefficient vectors $\mathbf{L}_i \in \mathbb{R}^{(s+1)^2}$. The network predicts per-Gaussian diffuse transport coefficients $\mathbf{T}_{k}$, and the diffuse color is \yr{calculated} by inner products in SH domain, modulated by the albedo $\rho_k$:
\begin{equation}
    c_{\text{diffuse}, k}(\mathcal{L}) = \rho_k \odot \sum_{i=1}^{M} \left( \mathbf{T}_{k} \cdot \mathbf{L}_i \right).
    \label{eq:diffuse}
\end{equation}

The view-dependent \textbf{specular} term $c_{\text{specular}, k}$ is modeled using a simplified Cook-Torrance microfacet model~\cite{cook1981reflectance} to create realistic highlights.
The total specular radiance is the sum of contributions from all $M$ light sources:
\begin{equation}
    c_{\text{specular}, k}(\mathcal{L}, \omega_o) = \sum_{i=1}^{M} f_{\text{specular}}(\omega_i, \omega_o, \mathbf{n}_k) (\mathbf{n}_k \cdot \omega_i) \mathbf{I}_i,
    \label{eq:specular_sum}
\end{equation}
where the specular BRDF $f_{\text{specular}}$ is defined as:

\begin{equation}
    f_{\text{specular}}(\omega_i, \omega_o, \mathbf{n}_k) = \frac{D(\mathbf{h}_i, \alpha_k) \, F(\omega_o, \mathbf{h}_i) \, G(\omega_i, \omega_o, \alpha_k)}{4 \, (\mathbf{n}_k \cdot \omega_i) \, (\mathbf{n}_k \cdot \omega_o)},
    \label{eq:cook-torrance}
\end{equation}

Here, $\mathbf{h}_i = \frac{\omega_i + \omega_o}{\|\omega_i + \omega_o\|}$ is the halfway vector for the $i$-th light source. The $D$ term is modeled by the Trowbridge-Reitz GGX distribution~\cite{trowbridge1975average} parameterized by the per-Gaussian roughness $\sigma_k$: 
\begin{equation}
    D(\mathbf{h}_i) = \frac{\alpha_k^2}{\pi ((\mathbf{n}_k \cdot \mathbf{h}_i)^2 (\alpha_k^2 - 1) + 1)^2}, \quad \alpha_k = \sigma_k^2.
    \label{eq:ggx-d}
\end{equation} 
The Fresnel term $F$ is computed using Schlick’s approximation~\cite{schlick1994inexpensive}
\begin{equation}
    F(\omega_o, \mathbf{h}_i) = F_0 + (1 - F_0)(1 - \max(\omega_o \cdot \mathbf{h}_i, 0))^5,
    \label{eq:fresnel}
\end{equation}
with a fixed dielectric Fresnel base reflectance $F_0 = 0.04$ for the entire head, consistent with skin and hair materials.  
The geometry term $G$ follows the Smith masking-shadowing model for GGX as described in~\cite{walter2007microfacet}.

\subsection{Disentanglement}
Our goal is to disentangle intrinsic head appearance and extrinsic illumination in a generative 3D Gaussian head model, supervised by high-fidelity lightstage captures. 
To achieve stable training, we devise a three-stage training strategy:
\yr{(1)} The first stage \yr{trains} the base branch $G_{base}$, adapting a pretrained entangled generator to light-invariant domain.
\yr{(2)} Based on the guidance of \yr{the} base branch, the second stage \yr{trains the} relight branch $G_{relit}$, learn\yr{ing} the interaction between incident light and head surfaces using the captured lightstage data. 
(3) To generalize our model to arbitrary illumination beyond the limited white-light OLAT data, we apply adversarial loss on images rendered with randomly sampled lighting.

\subsubsection{\yr{Stage 1:} Base Branch Adaptation}
We initialize $G_{base}$ by adapting a pretrained head generative model\yr{, GGHead}~\cite{gghead} to produce $\mathcal{A}_{\text{geo}}$ and view-independent albedo $\rho_k$, while maintaining its diversity of high-quality generation.
\yr{GGHead does not decouple intrinsic appearance and scene illumination, and its color}
output \yr{--} first-order SH coefficients per RGB channel \yr{--} represents \yr{a} view-dependent radiance \yr{that entangles both factors}. 
We firstly modify GGHead to output 3-channel RGB colors instead of the original 12-channel SH coefficients. 
The modified generator is first finetuned on the FFHQ dataset (with segmentation masks derived from SAM~\cite{kirillov2023segment} and \cite{zheng2024bilateral}) using \yr{an} adversarial loss, \yr{which} enforces a view-independent radiance, discarding anisotropic shading components at the base level. 
Subsequently, it is finetuned for 20k iterations using pseudo albedo images generated by~\cite{nerffacelighting} via adversarial loss. 
Instead of updating all layers, we restrict updates to the intermediate layers (specifically, layer 5 and 6 in the GGHEAD backbone).
These intermediate layers are known to encode mid-level appearance features such as shading and illumination~\cite{karras2019style}.  
This targeted adaptation steers the generator toward the albedo domain while preserving high-level identity structure and face details, and the limited training duration further mitigates the risk of collapsing to the limited pseudo albedo images.

\subsubsection{\yr{Stage 2:} Relight Priors from LightStage Captures}
Stage 2 focuses on \yr{training} the relight branch $G_{\text{relit}}$\yr{, while keeping} the base generator $G_{\text{base}}$ frozen. 
We first implement multi-view \yr{GAN} inversion \yr{on $G_{base}$} to obtain geometry and albedo of lightstage captures. 
\yr{We then} optimize $G_{relit}$ and latent code per subject to predict relighting-related attributes $(\mathbf{n}_k, \sigma_k, \mathbf{T}_{d,k})$ \yr{that} match the lightstage observations, \yr{where} the multi-scale features \yr{output by} $G_{base}$ \yr{are injected into $G_{\text{relit}}$ as a guidance}. 
A novel normal distillation strategy is proposed to predict high-quality normal. 
We further utilize several regularization terms to constrain the generated attributes to be plausible.

\noindent\textbf{Geometry and Albedo Inversion.} 
Given an image sequence captured under varying lighting, we perform multi-view \yr{GAN} inversion on the fully illuminated frame and propagate the results to the following frames.
Specifically, we optimize implicit latent code $\mathbf{w}^+ \in \mathbb{R}^{512 \times L}$ in the extended W$^+$ space of StyleGAN2 
to yield 
per-Gaussian attributes $\mathcal{A}_{\text{geo}}$ and albedo $\rho_k$, as well as intermediate feature maps $\{\mathbf{F}^l_{\text{base}}\}_{l=1}^L$ from $G_{\text{base}}$.
We use L2 loss and perceptual loss to constrain the multi-view inversion, and image total variance loss to avoid holes \yr{during} gaussian rendering:
\begin{equation}
    \mathcal{L}_{inv} = \mathcal{L}_{2} + \mathcal{L}_{perc} + \mathcal{L}_{tv}.
\end{equation}
Notably, unlike methods such as PTI~\cite{roich2022pivotal} that fine-tune the generator weights per subject, our approach keeps the base generator \( G_{\text{base}} \) fixed throughout inversion to preserve consistent feature representations across different identities in the latent space.

\noindent\textbf{Multi-Scale Feature Guidance.}
Observing that early layers of $G_{\text{base}}$ encode coarse shape (informing global illumination effects), while later layers capture fine details (crucial for normals and roughness), we leverage multi-scale features $\{\mathbf{F}^l_{\text{base}}\}_{l=1}^L$ of $G_{base}$ to guide the relight branch. 
Specifically, a lightweight fusion \yr{network} $\Phi^l$ \yr{predicts} a residual update $\Delta^l$ from the concatenation of the current features of relight branch $\mathbf{F}'^l_{\text{relit}}$ and the instance-normalized base branch feature $\mathbf{F}^l_{\text{base}}$:
\begin{equation}
    \mathbf{F}^l_{\text{relit}} = \mathbf{F}'^l_{\text{relit}} + \Delta^l, 
    \quad \Delta^l = \Phi^l\big(\mathbf{F}^l_{\text{base}}, \mathbf{F}'^l_{\text{relit}}\big).
    \label{eq:fusion}
\end{equation}
This \yr{guidance} enables $G_{\text{relit}}$ to predict relighting attributes consistent with the underlying geometry and albedo from $G_{\text{base}}$ while ensuring robust training.

\noindent\textbf{Normal Prediction.}
Notably, 3DGS does not provide native normals, which are critical for view-dependent specular rendering. 
One viable method is to extract an explicit mesh from the 3D Gaussian splats and assign nearest face normal to each splat. But mesh extraction for each training iteration is computationally expensive, and the recovered surfaces are sensitive to noise of gaussian splats. 
An alternative is to use k-NN search to find neighboring splats and compute average orientation as the normal, which is even more susceptible to noise.
\yr{Since} our $G_{base}$ module generates high-quality geometry, its output features $\{\mathbf{F}^l_{\text{base}}\}_{l=1}^L$ implicitly contain the geometry information\yr{, which could serve as cues for predicting normals}. 
\yr{As we have injected the multi-scale base branch feature $\{\mathbf{F}^l_{\text{base}}\}_{l=1}^L$ into the relight branch,} we \yr{propose} to directly \yr{use} $G_{relit}$ \yr{to} predict the normals\yr{, adding 3 channels to its output as the normal map}. 

To ensure the correctness and smoothness of the predicted normals, we introduce a self-distillation loss and 2D smoothness term.
\yr{Based on} the inverted geometry represented by gaussian point clouds, we pre-extract meshes via poisson reconstruction and cleanup the noise hair region by projecting onto 2D skin-parsing~\cite{face-parsing} masks.
Let $\hat{\mathbf{n}}^k_{\text{mesh}}$ be the closest mesh triangle normal of the gaussian splat $g_k$. 
We minimize the angular deviation between normalized normal prediction $\mathbf{n}_k$ and $\hat{\mathbf{n}}_k^{\text{mesh}}$ via cosine similarity:
\begin{equation}
        \mathcal{L}_{normal}^{D} = \frac{1}{N} \sum_{k=1}^{N}  (1 -  \mathbf{n}_k^\top \hat{\mathbf{n}}_k^{\text{mesh}}).
\end{equation}

For the noisy hair region, we utilize a 2D \textbf{T}otal \textbf{V}ariation loss.
We render the normalized normals $\mathbf{n}_k$ as RGB colors of gaussian and splat them to a 2D normal map $I_{\text{norm}} \in \mathbb{R}^{H \times W \times 3}$, and apply total variation regularization:
\begin{equation}
        \mathcal{L}_{normal}^{TV} = \left\| \nabla_x I_{\text{norm}} \right\|_1 + \left\| \nabla_y I_{\text{norm}} \right\|_1,
\end{equation}
where $\nabla_x$ and $\nabla_y$ denote horizontal and vertical finite differences.

\noindent\textbf{Training.}
The final loss of Stage 2 consists of shading loss, the aforementioned normal prediction loss and a regularization term on diffuse transport cofficients $T_k$.
\yr{We use these losses} to optimize parameters of $G_{relit}$ and a latent code $\mathbf{w}^+_{lit} \in \mathbb{R}^{512 \times L}$ to represent per-subject head appearance properties.

The shading loss is \yr{formulated} in Eq.~(\ref{eq:shade}), where $\mathcal{L}_{tv}^I$ is involved to prevent image noise.
\begin{equation}
    \mathcal{L}_{shading} = \mathcal{L}_{\text{img}} + \mathcal{L}_{perc} + \mathcal{L}_{tv}^I.
    \label{eq:shade}
\end{equation} 

To encourage spectrally smooth diffuse response, we regularize the diffuse transport coefficients $\mathbf{T}_{k} \in \mathbb{R}^{3 \times 9}$ by minimizing the variance across the RGB channels:
\begin{equation}
    \mathcal{L}_{\text{prt}} = \frac{1}{N} \sum_{k=1}^{N} \| \mathbf{T}_{k} - \bar{\mathbf{T}}_{k}\|_2^2, \quad \bar{\mathbf{T}}_{k} = \frac{1}{3} \sum_{c \in \{r,g,b\}} \mathbf{T}_{k}^{(c)}. 
\end{equation}

\begin{figure*}[t]
    \centering
    \includegraphics[width=0.95\textwidth]{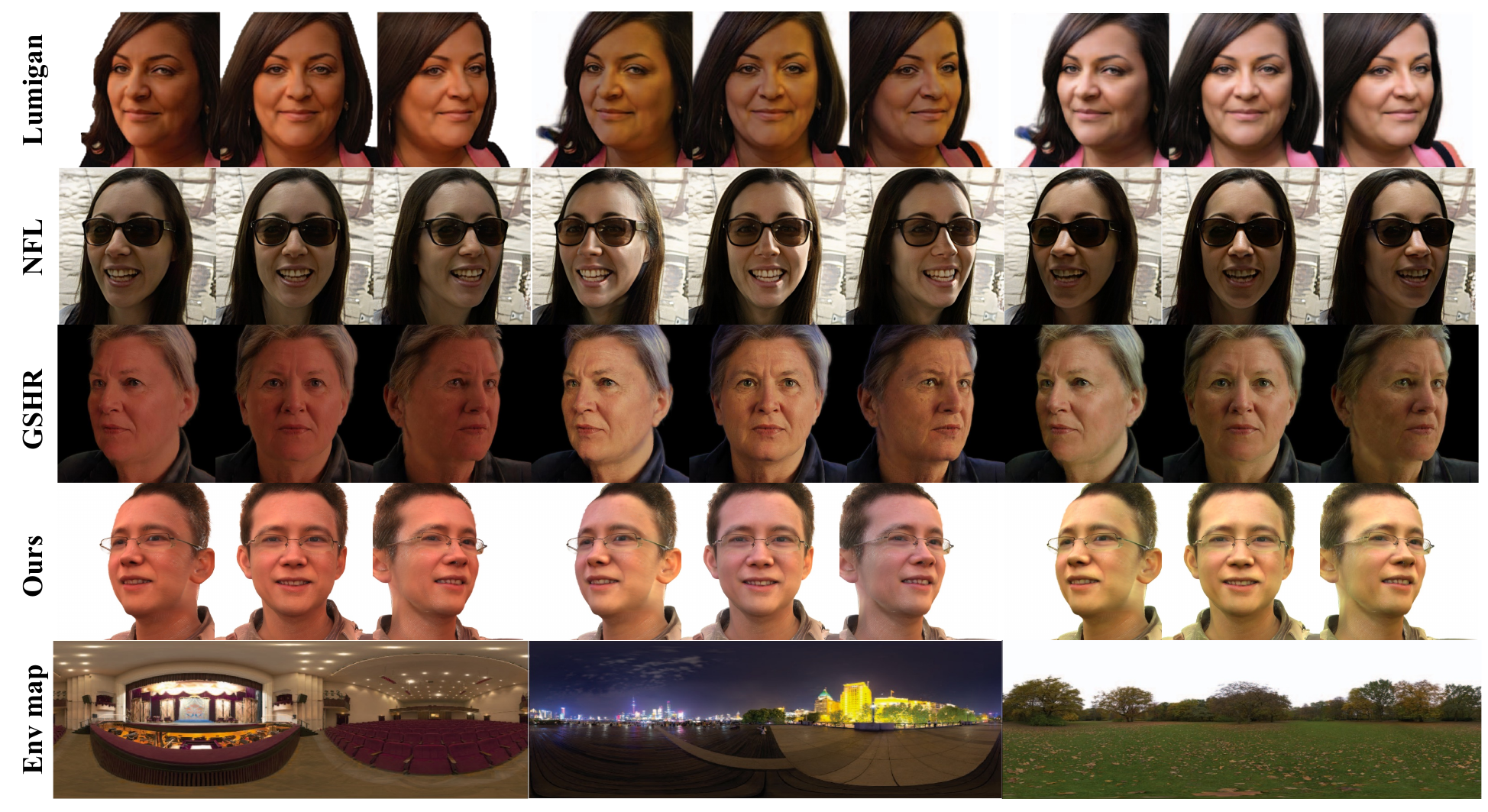}
    \vspace{-10pt}
    \caption{Relighting under complex environment maps. LuminGAN~\cite{deng2024lumigan} and NFL assumes white illumination and fails to handle colored lighting. GSHR lacks real-world OLAT supervision and therefore cannot produce correct responses under non-uniform colored lighting. In contrast, our method accurately reproduces spatially varying lighting effects.}
    \label{fig:envmap}
\end{figure*}

\subsubsection{Stage3: Adversial Training}
In this final stage, we perform adversial training to ensure robust photorealism under arbitrary illumination.
Given that our dataset consists solely of limited white-light OLAT captures, we employ adversarial supervision on images generated with randomly sampled illumination. This strategy expands the training distribution, ensuring the model generalizes well to diverse illumination environments.

First, to support random sampling of the relight latent code, we initialize mapping network $M_{\text{lit}}$ that maps a randomly sampled code $z_{lit} \in \mathbb{R}^{128}$ into the $\mathcal{W}^+$ space of $G_{relit}$, utilizing the optimized $\mathbf{w}^+_{\text{lit}}$ from Stage 2 as training data. 
Subsequently, we freeze the parameters of the base branch $G_{base}$, exclusively optimizing the synthesis network of $G_{relit}$.
During each training iteration, we simulate diverse illumination conditions to broaden the lighting manifold beyond the training set. 
We pre-define a set of candidate point lights uniformly distributed on the hemisphere frontal to the face. 
A random lighting condition $\mathcal{L}_{rand}$ is constructed by randomly sampling a subset of candidate lights and assigning random RGB intensities to each source. 
Simultaneously, we sample random latent codes $\mathbf{z}_{base}$ for the geometry representation and \yrn{random} $\mathbf{z}_{lit}$ for the materials. 
The relight branch $G_{relit}$ synthesizes the relighted head image using these random inputs, which is then supervised by an adversarial loss using pretrained discriminator of GGHead. 
This supervision guides the generator to produce high-fidelity, photorealistic textures that remain stable under varying, colored synthetic lighting. 
Moreover, the losses from Stage 2 are retained in this final stage to serve as a regularizer to prevent the relight branch from drifting.
To ensure stable fine-tuning, we employ a reduced learning rate of $1 \times 10^{-6}$.
\section{Experiments}
\label{sec:exp}

\begin{table}
\centering
    \begin{tabular}{ccccc}
    \toprule
        Method & SR & Relit & FID$\downarrow$ & FPS$\uparrow$ \\
    \midrule
         EG3D& \checkmark & $\times$ & 4.30 & 42 \\
         GGHead& $\times$ & $\times$ & 4.87 & 228\\
    \midrule
         NFL& \checkmark & \checkmark &4.16 & 2.8 \\
         GSHR & $\times$ & \checkmark &5.71 & 243\\
    \midrule
         Ours & $\times$& \checkmark & 4.78 & 201\\
    \bottomrule
    \end{tabular}
    \caption{Comparison of generative quality and rendering speed. \textbf{SR} denotes super resolution modules and \textbf{Relit} denotes whether to support explicit lighting control. Rendering speed is tested on Nvidia A6000 GPU.}
    \label{tab:generation}
    \vspace{-2.5em}
\end{table}

\subsection{Implementation Details.}
\label{sec:dataset}
\noindent\textbf{Lightstage Dataset.} We first perform calibration to obtain the world coordinates of each point light and camera.
For the captured multi-view images, we use open-source tools to detect 2D landmarks~\cite{bulat2017far} and human segmentation~\cite{zheng2024bilateral} for the "Uniform" frame. 
All the frames are aligned and cropped into 512 resolution.
The head pose for each frame is solved using multi-view 2D landmarks and calibrated camera extrinsics, constrained by a 3DMM prior~\cite{paysan20093d}.
$56$ participants (31 male, 25 female) perform three expressions (neutral, smile, and eyes closed), each recorded under $95$ distinct lighting configurations. We select $46$ subjects for training and $10$ for testing.

\noindent\textbf{Architecture and Training.}
\( G_{\text{relit}} \) shares the same overall architecture and number of layers as the \( G_{\text{base}} \). 
However, since \( G_{\text{base}} \) models complex geometric and appearance details that go far beyond head reflectance alone, we design \( G_{\text{relit}} \) with reduced capacity: the convolutional layers in \( G_{\text{relit}} \) operate on features of up to 256 dimensions, compared to up to 512 dimensions in \( G_{\text{base}} \). 
Correspondingly, we use a 128-dimensional latent code \( \mathbf{z}_{\text{lit}} \) to represent head material properties, while geometry and intrinsic appearance are encoded by a 512-dimensional latent code \( \mathbf{z}_{\text{base}} \).
During Stage 2, \( G_{\text{relit}} \) is optimized with Adam using a learning rate of \( 5 \times 10^{-4} \), while the lighting latent code \( \mathbf{w}_{\text{lit}} \) is updated with a higher learning rate of \( 1 \times 10^{-3} \). 
Additional details are provided in the supplementary material.

\subsection{3D Generation.}
We evaluate generation quality against multiple state-of-the-art 3D-aware head synthesis methods.
First, we assess \textit{generation diversity} by generating 50K images and computing FID against the FFHQ dataset~\cite{kazemi2014one}. Images are rendered under random lighting conditions synthesized as described in Stage 3. 
Second, we measure \textit{rendering speed} on a NVIDIA A6000 GPU. 
Third, we evaluate \textit{multi-view spatial consistency}: for a random latent code, we render 30 viewpoints across $-45\degree$ to $45\degree$, hold out the frontal view as ground truth to calculate metrics, and use the remaining views to reconstruct the geometry by NeuS2~\cite{wang2023neus2}. 

We compare with four baselines: EG3D~\cite{eg3d} and NFL~\cite{nerffacelighting}, which rely on neural volume rendering, and GGHead~\cite{gghead}, GSHR~\cite{lv2025gsheadrelight}, which are built upon 3D Gaussian splatting. 
Among them, EG3D and GGHead do not support explicit illumination control while NFL and GSHR support lighting control.
We show the FID score and rendering FPS in Table~\ref{tab:generation}.
Methods based on 3D Gaussian achieve significantly higher rendering speeds than volume-rendering counterparts. Our method is slightly slower than GGHead due to the physically-based shading model but maintains the real-time rendering performance.
We evaluate multi-view consistency using a wide viewport range (±45°).
Our method achieves plausible disentanglement, which eliminates the inconsistent anisotropic artifacts, thus produces better multi-view consistency, as shown in Table~\ref{tab:consistent}.

\begin{table}
\centering
    \begin{tabular}{cccc}
    \toprule
        Method & PSNR $\uparrow$ & SSIM$\uparrow$ & LPIPS$\downarrow$ \\
    \midrule
         EG3D& 29.29 & 0.9026 &  0.065\\
         GGHead& 31.53& 0.9085 & 0.052\\
         NFL& 30.02 & 0.9002  & 0.058\\
         GSHR & 30.64 & 0.8890  & 0.062\\
         \midrule
         Ours & \textbf{32.09} & \textbf{0.9132} & \textbf{0.050} \\
    \bottomrule
    \end{tabular}
    \caption{Evaluation of spatial consistency of generated 3D heads. We render multi-view images from the generated 3D heads and synthesis novel view using NeuS2~\cite{wang2023neus2}; higher PSNR, SSIM, and lower LPIPS indicate better spatial consistency.}
    \label{tab:consistent}
    
\end{table}

\subsection{Light Control.}
\begin{figure*}[t]
    \centering
    \includegraphics[width=0.96\textwidth]{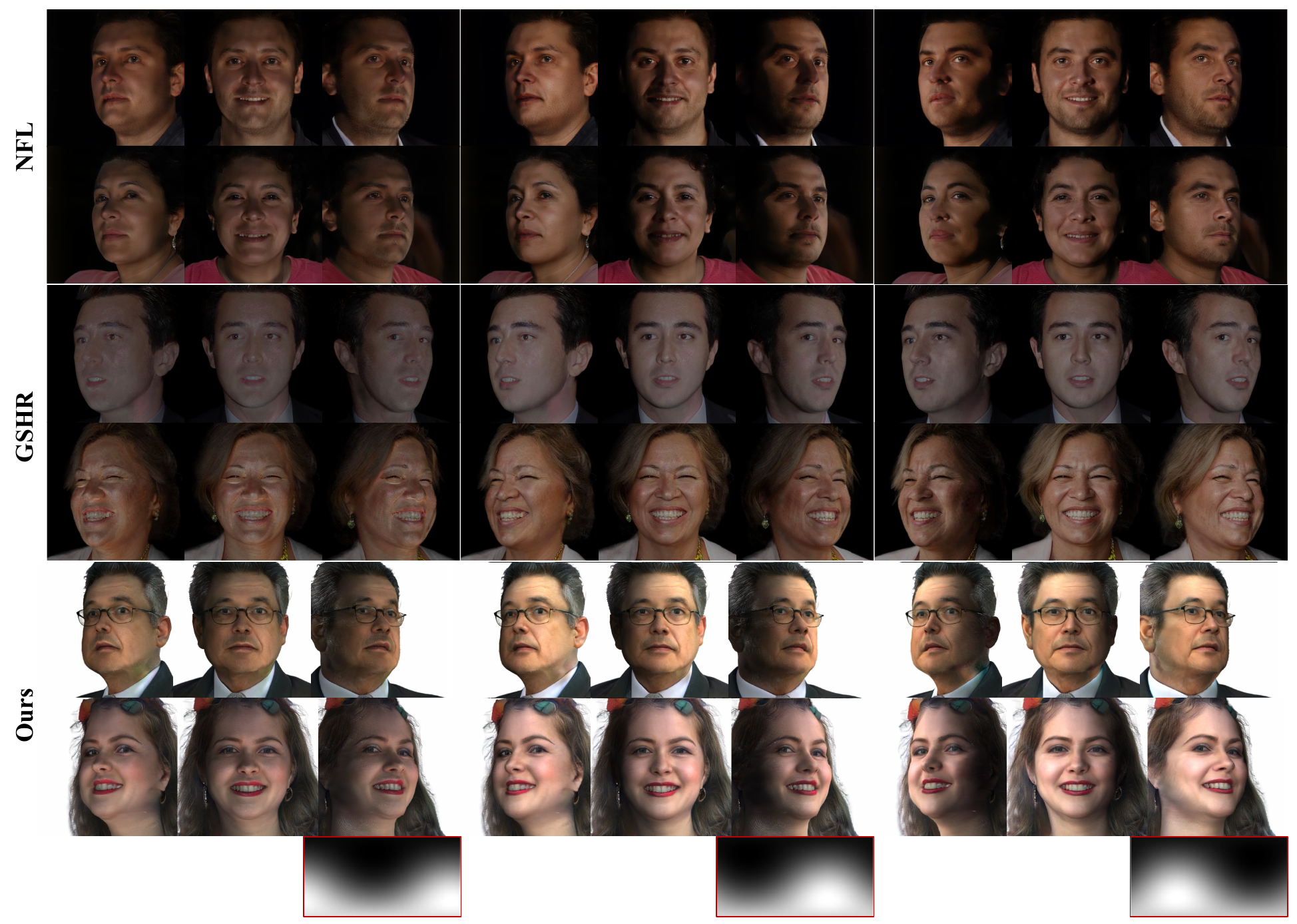}
    \caption{Relighting under complex environment maps. NFL assumes white illumination and fails to handle colored lighting. GSHR lacks real-world OLAT supervision and therefore cannot produce physically plausible responses under complex lighting. In contrast, our method, trained with multi-view OLAT captures, accurately renders realistic shading and reflections under diverse and challenging illumination conditions.}
    \label{fig:sh}
\end{figure*}

We compare the rendering performance of various generative methods under complex and extreme illumination conditions. 
The baselines include LumiGAN~\cite{deng2024lumigan}, NFL~\cite{nerffacelighting}, and GSHR~\cite{lv2025gsheadrelight}. 
While LumiGAN and NFL are restricted by a white-light assumption , both GSHR and our method support colored illumination. Since LumiGAN’s code is unavailable, its results are sourced from the original paper.
Figure~\ref{fig:envmap} illustrates the results under non-uniform colored environment maps. Our generated 3D heads accurately respond to spatially-varying environmental lighting. 
Furthermore, Figure~\ref{fig:sh} demonstrates relighting results under extreme conditions, such as side-lighting or low-light environments. Lacking supervision from real-world data, NFL and GSHR fail to handle these extreme cases, either producing significant artifacts or failing to respond. In contrast, our method produces high-fidelity shadow and specular highlights.

\subsection{Portrait Relighting.}
\begin{figure*}[t]
    \includegraphics[width=0.94\textwidth]{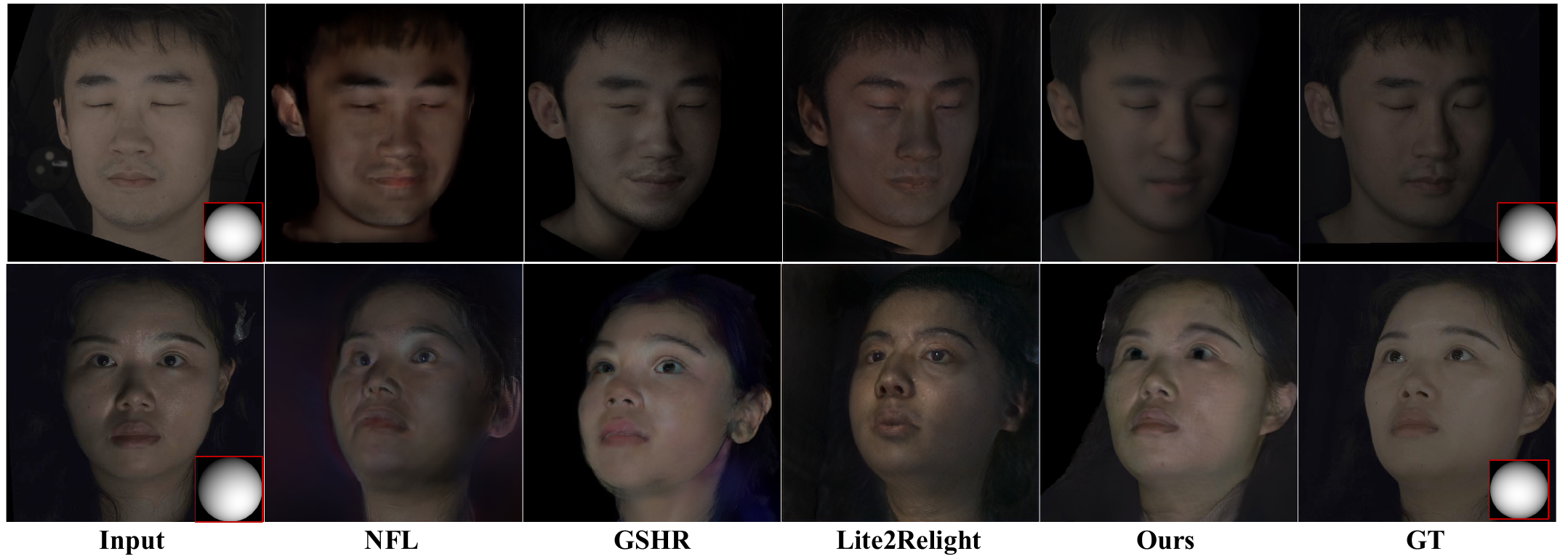}
    \caption{Free-view relighting on captured LightStage data. Compared with NFL, GSHR, and Lite2Relight, our method better disentangles lighting from intrinsic facial attributes by learning reflectance priors from OLAT data, leading to more realistic relighting from a single portrait image.
    We show the lighting condition of the image in the red boxes.}
    \label{fig:lightstage}
\end{figure*}

Our method, along with NFL and GSHR, enable generating 3D human heads and support portrait illumination and viewport editing via 3D GAN inversion.
To validate our model’s ability to disentangle illumination and appearance, we extract image pairs from LightStage captures and perform free-view relighting using our method and baselines.
We also compare against Lite2Relight\cite{lite2relight}, a recent 3D GAN based portrait editing method.
For 10 test subjects, we randomly sample two types of OLAT image pairs: one from uniform illumination to side lighting (as shown in the first row of Fig~\ref{fig:lightstage}), and the other from side lighting to uniform illumination (shown in the second row of Fig~\ref{fig:lightstage}), with 10 random lighting combinations for each type.
Our method demonstrates superior robustness in handling low-light environments and extreme side-lighting.
We also provide quantitative results in Tab~\ref{tab:consistent}.

\begin{table}[t]
\centering
    \begin{tabular}{cccc}
    \toprule
        Method & PSNR $\uparrow$ & SSIM$\uparrow$ & LPIPS$\downarrow$ \\
    \midrule
         NFL& 24.6011 & 0.8227 & 0.1179 \\
         GSHR & 20.0688 & 0.7464 & 0.1907 \\
         Lite2Relight & 20.8874 & 0.8073 & 0.1278\\
         \midrule
         Ours & \textbf{27.0550} & \textbf{0.8614} & \textbf{0.1001} \\
    \bottomrule
    \end{tabular}
    \caption{Quantitative comparisons of free view relighting.}
    \label{tab:consistent}
    \vspace{-2.5em}
\end{table}

\subsection{Ablation Study.}
\begin{figure}[t]
    \includegraphics[width=0.95\linewidth]{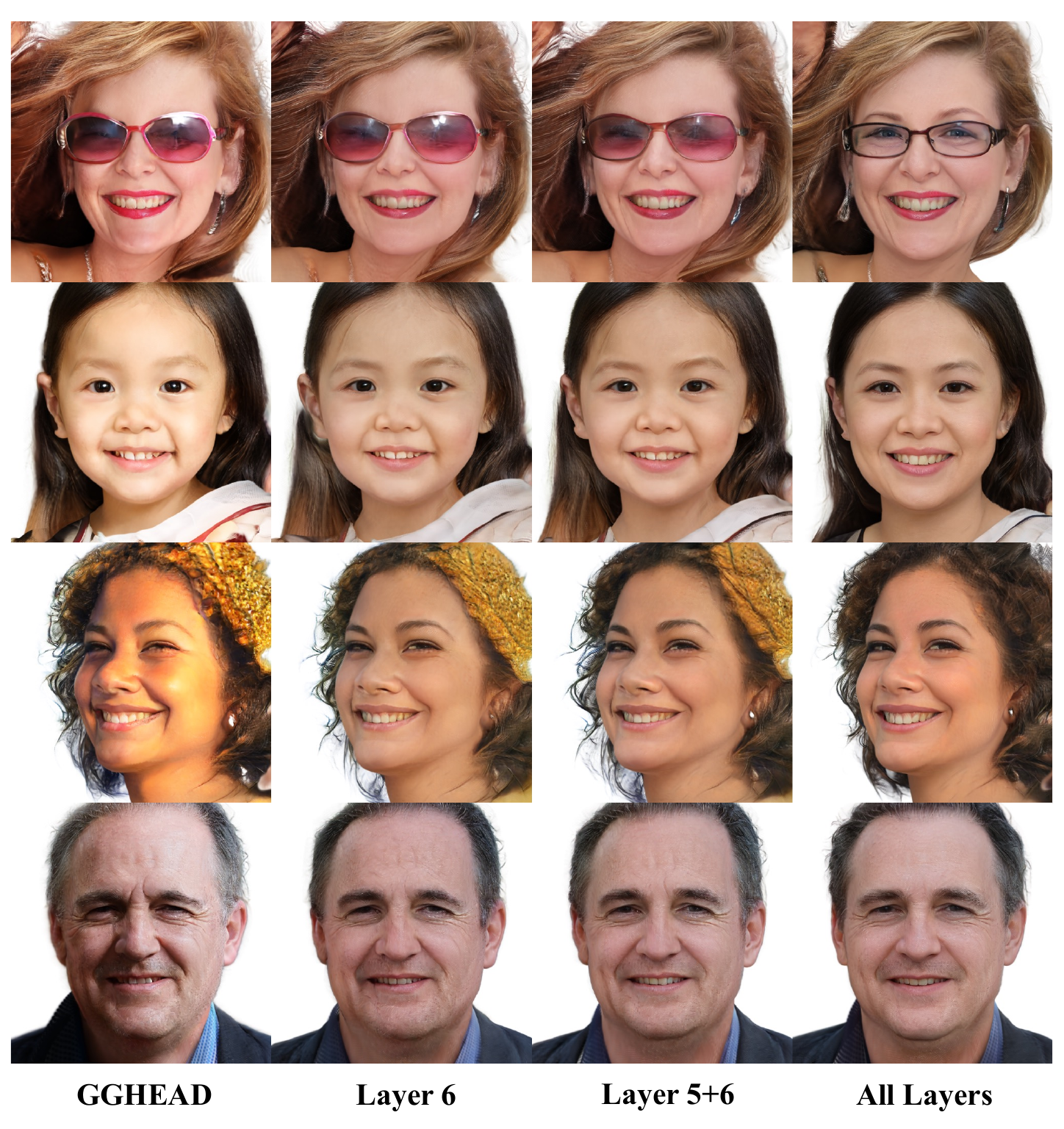}
    \caption{Effect of optimizing different layers for illumination removal. From left to right: original GGHead output, optimization of layer 6 only, layers 5+6, and full-model optimization. Optimizing only layer 6 fails to remove all lighting artifacts (e.g., sunglasses in row 1, left mouth corner in row 4). Full-model optimization significantly alters facial geometry (e.g., rows 2–3).}
    \label{fig:albedo_finetune}
    \vspace{-1em}
\end{figure}

\noindent\textbf{Base Branch Adaption.}
We fine-tune GGHead with NFL-generated pseudo-albedo images to remove illumination, optimizing only intermediate layers to avoid identity collapse. 
We compare results before and after optimization using the same latent code, as shown in Fig~\ref{fig:albedo_finetune}. From left to right: the original GGHead output, optimization of layer 6, optimization of two intermediate layers 5 and 6, and full-model optimization. Full-model optimization significantly alters facial geometry, biasing it toward the average face. Optimizing only the 6th layer fails to fully remove illumination artifacts. Our method effectively removes illumination while preserving coarse face structure and fine-grained details.

\begin{figure}[t]
    \begin{center}
    \includegraphics[width=\linewidth]{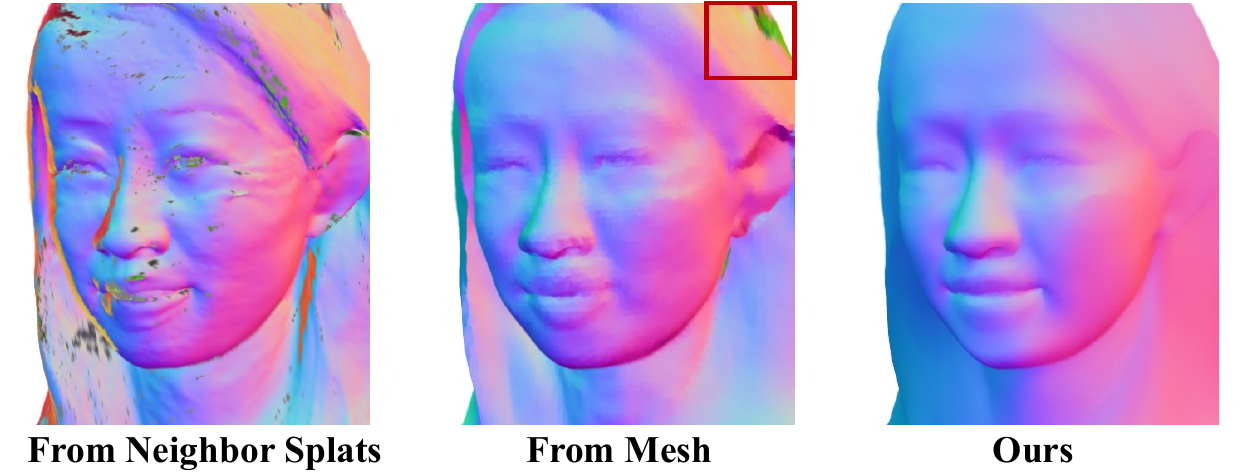}
    \caption{Normal generation using different methods. From left to right: normals from averaged orientations of neighboring Gaussians, normals computed after mesh extraction, and our method. The first two approaches exhibit noticeable noise (highlighted in red boxes), while our method produces plausible and smooth normals.}
    \label{fig:normal}
    \end{center}
    \vspace{-1em}
\end{figure}

\noindent\textbf{Normal Distillation.}
We show the normal calculation results of different methods in Fig~\ref{fig:normal}.
Directly computing normals from neighboring points in gaussian splats introduces significant noise, making it unsuitable for photorealistic rendering. 
Reconstructing a mesh from the Gaussian point cloud and then extracting normals is computationally expensive to the training process and often produces noticeable artifacts in fine regions such as hair. 
To address this, we directly predict normals using the $G_{relit}$ branch, supervised by normals derived from a pre-extracted mesh combined with a 2D total variation loss, yielding smooth and accurate surface normals.

\begin{figure}[t]
    \begin{center}
    \includegraphics[width=\linewidth]{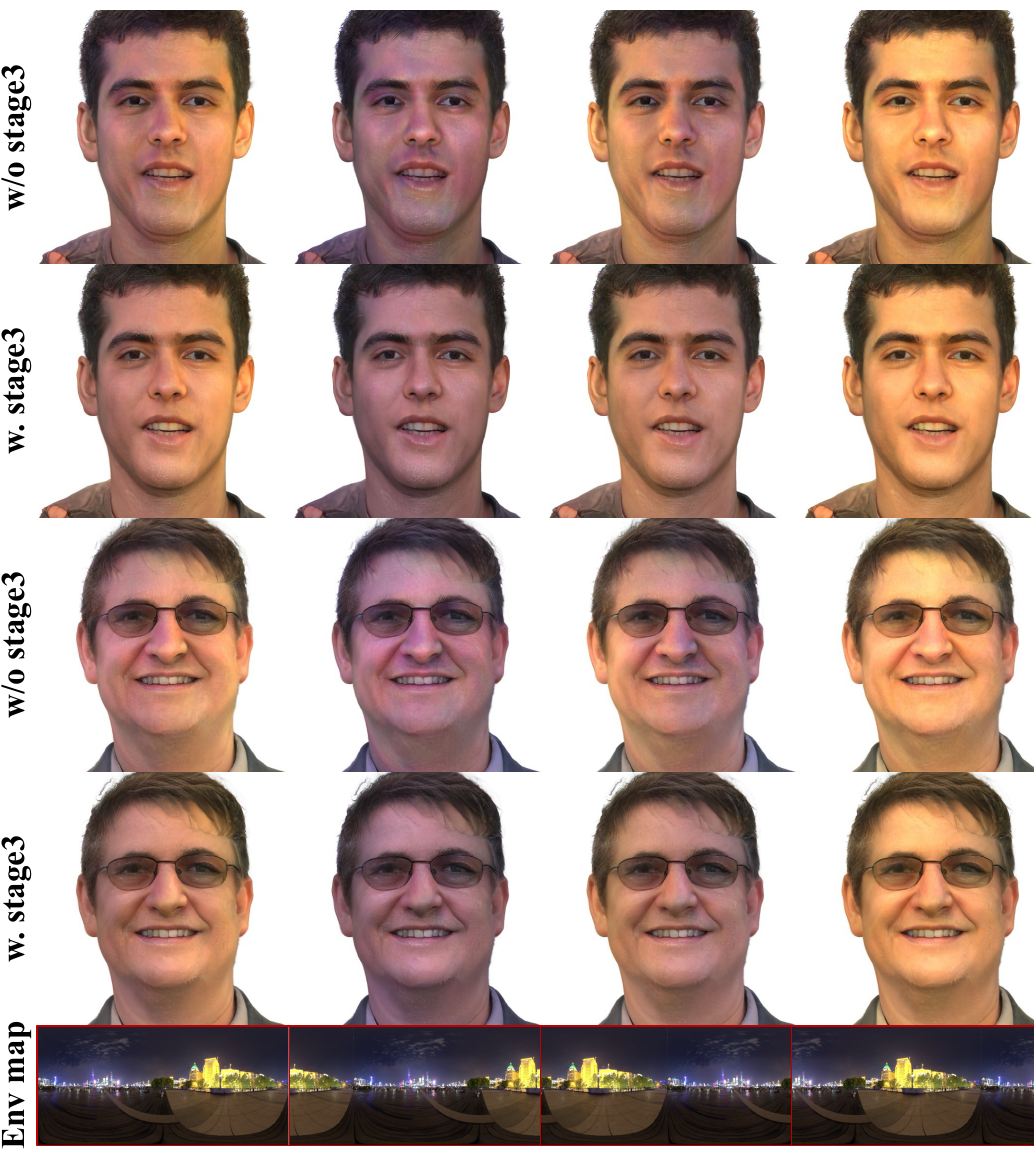}
    \caption{Ablation study on Stage 3. We rotate the environment map and render two randomly sampled 3D heads to evaluate the impact of Stage 3 adversarial training. Rows 1 and 3 show results without Stage 3 training. Under blue lighting—an illumination condition significantly different from the OLAT training distribution—models without Stage 3 exhibit noticeable artifacts. In contrast, Stage 3 training yields more plausible and robust relighting results.}
    \label{fig:joint}
    \end{center}
    \vspace{-1em}
\end{figure}

\noindent\textbf{Random Light and Adversarial Training.}
We evaluate the impact of random illumination and adversarial training by comparing rendering results under novel, colored lighting conditions not present in the white-light OLAT training set.
As shown in Fig~\ref{fig:joint}, without this stage, the model exhibits noticeable artifacts and color inconsistencies when subjected to non-uniform illumination.
By incorporating adversarial training with randomly sampled lighting, our method ensures well generalization to diverse lighting, producing stable and realistic relighting results under challenging, unseen conditions.

\begin{figure}[t]
    \begin{center}
    \includegraphics[width=0.8\linewidth]{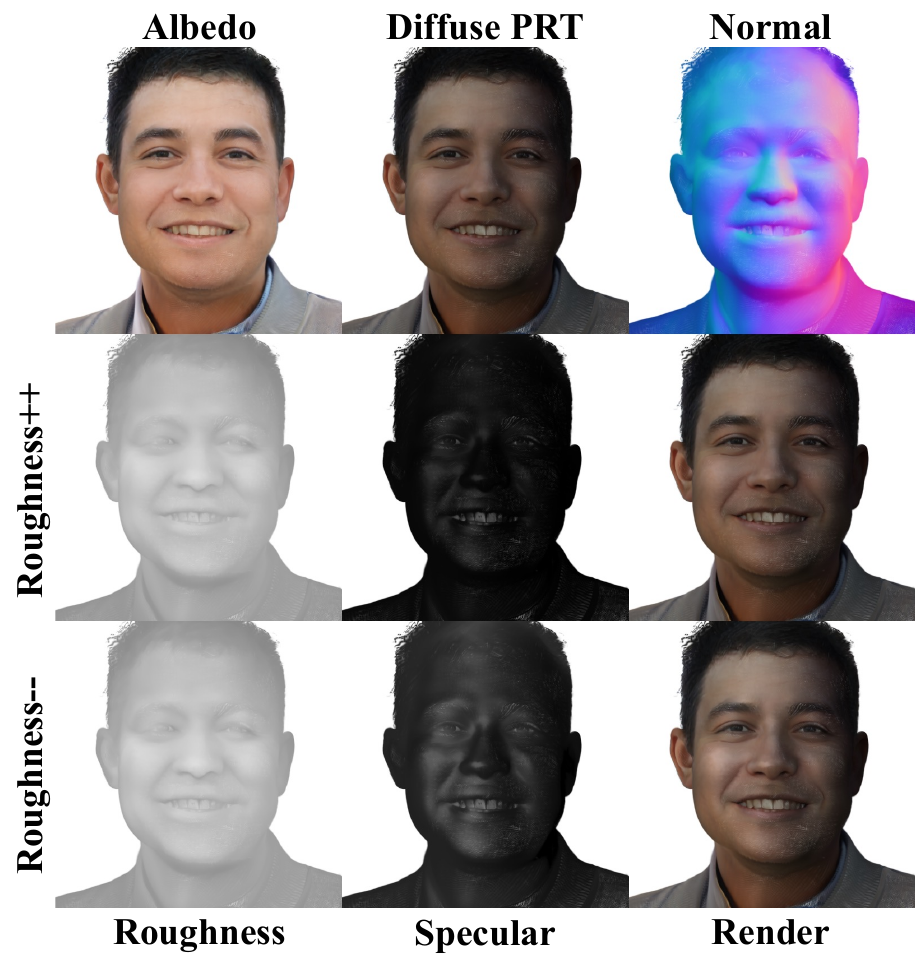}
    \caption{Explicit roughness editing. We demonstrate the editing of material roughness for the generated characters. By decreasing the roughness value (i.e., increasing smoothness), we achieve more pronounced specular highlights, as shown in the third row. Notably, the underlying geometry and albedo remain unchanged, highlighting the disentangled nature of our representation.}
    \label{fig:ws_lit}
    \end{center}
    \vspace{-1em}
\end{figure}

\noindent\textbf{Explicit Control on Roughness.}
Our method generates disentangled and physically-based face material properties. To evaluate the quality of disentanglement, we perform direct attribute editing on the roughness component. As illustrated in Figure~\ref{fig:ws_lit}, decreasing the roughness value leads to more pronounced specular highlights while preserving other appearance attributes, demonstrating effective decoupling of the material properties.

\section{Conclusion.}

We propose a novel framework to disentangle illuminations and intrinsic appearances within a 3D gaussian head generative model supervised by high-fidelity lightstage captures.
We utilize a dual-branch architecture to predict geometry with albedo and shading attributes, and employ a progressive training schema to ensure stable training. We further propose a normal supervision method to produce high-quality normals which are crucial for realistic rendering. 
Experiments demonstrate the generation ability and rendering speed of our method, validate the quality of disentanglement by light and viewport control. 

\bibliographystyle{ACM-Reference-Format}
\bibliography{main}

\newpage

\clearpage

\clearpage

\newpage
\setcounter{page}{1}

\section{Implementation Details}
\begin{figure}[t!]
    \centering
    \includegraphics[width=\linewidth]{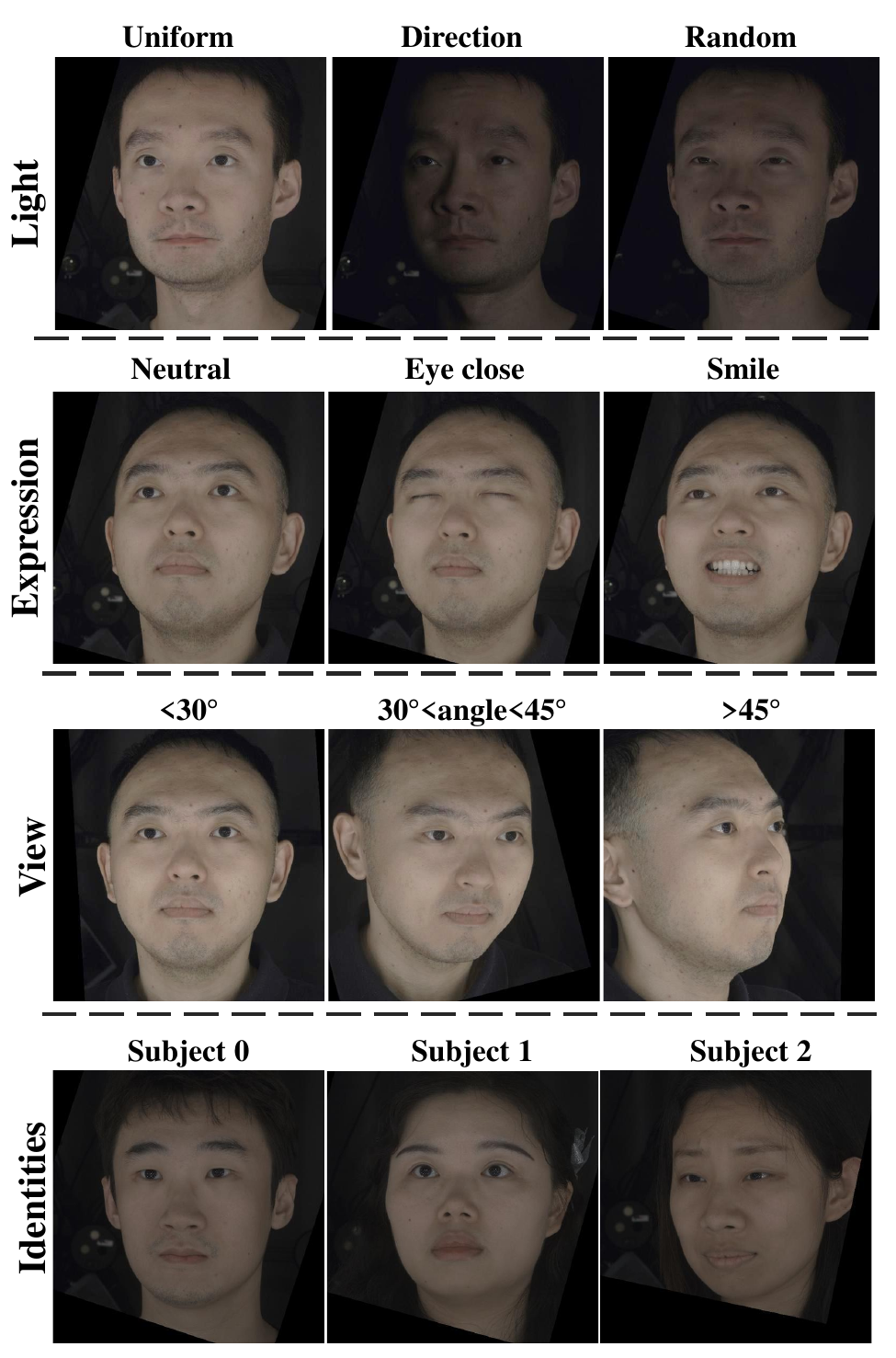}
    \caption{\textbf{An overview of the lightstage captures}. }
    \vspace{-1.5em}
    \label{fig:dataset}
\end{figure}

\begin{figure}[t!]
    \centering
    \includegraphics[width=\linewidth]{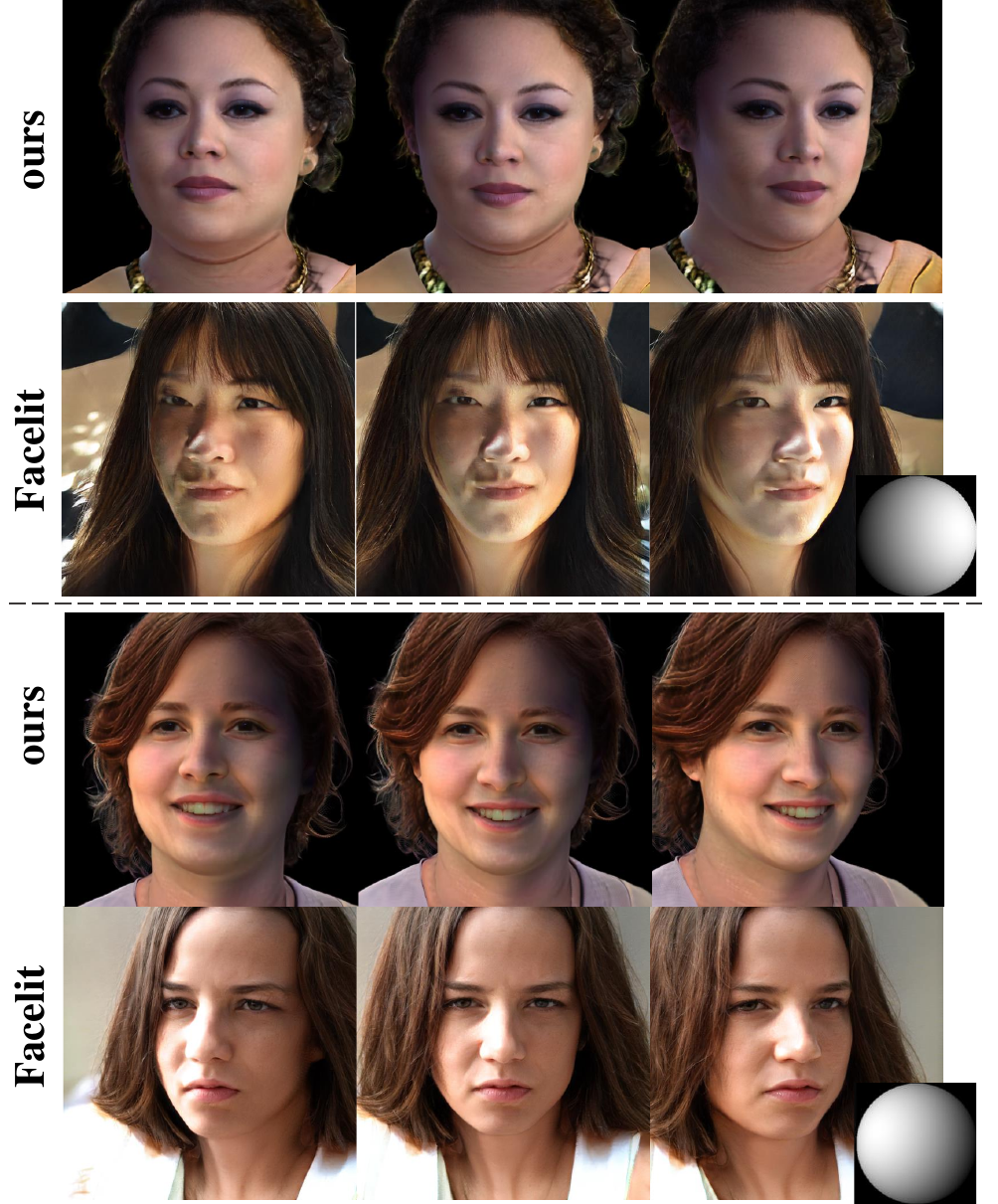}
    \caption{\textbf{Comparisons with FaceLit~\cite{ranjan2023facelit}}.}
    \label{fig:facelit}
    \vspace{-2.5em}
\end{figure}

\noindent\textbf{Dataset.}
We provide a summarized overview of the LightStage dataset in Fig~\ref{fig:dataset}. The dataset encompasses diverse subjects, lighting conditions, facial expressions, and viewing angles. The lighting conditions are categorized into three types: uniform, directional, and random. The facial expressions include neutral, smiling, and closed eyes.

\noindent\textbf{Training.} We provide hyper parameters and configurations of training in this section. 
\begin{itemize}
    \item \textbf{Stage1}: We retain the majority of the pretrained GGHead model’s architecture and parameters, and modify only the final layer to output RGB values instead of the original spherical harmonic coefficients. We fine-tune this modified model for 200k iterations on the FFHQ dataset using an adversarial loss. We then generate 10,000 pseudo-albedo images using NFL~\cite{nerffacelighting}. To prevent identity collapse, we use a large truncated $\psi$ value, which controls the diversity of sampled latent codes by constraining the sampling region in the latent space. We fine-tune the model for an additional 10,000 iterations on these pseudo-albedo images with an adversarial loss, removing the lighting component from the pretrained GGHead model while maintaining the diversity of the generated identities.
    \item \textbf{Stage2-Inversion}: We achieve multi-view inversion under the supervision of multiple loss terms. We set the weight of the $\mathcal{L}_2$ to 0.1, the weight of the $\mathcal{L}_{perc}$ to 1, and the weight of $\mathcal{L}_{tv}$ to 0.01. We compute the perceptual loss only using frontal images. The inversion process requires 500 iterations, and the inversion result of the previous fully lit frame serves as the initial value for the next frame to accelerate convergence.
    \item \textbf{Network}: We employ a lightweight network including 2 layers and Leaky ReLU activation to enable multi-scale features of $G_{base}$ to guide the relight branch. We design \( G_{\text{relit}} \) with a reduced feature dimensionality: its convolutional layers operate on 256-dimensional features, compared to 512 in \( G_{\text{base}} \). Finally, the relight branch requires approximately 90 MB memory storage.
    \item \textbf{Stage2-Training}: We set weights of $\mathcal{L}_{normal}^{D}$, $\mathcal{L}_{normal}^{TV}$ to be 1 and 10 respectively. The weights of $\mathcal{L}_{img}$, $\mathcal{L}_{perc}$ and $\mathcal{L}_{tv}^{I}$ are $0.1, 1, 1$. And weight of $\mathcal{L}_{prt}$ is set to $10$.
    \item \textbf{Stage3-Training}: Firstly, we train mapping network $M_{lit}$ for 10k iterations. Then we train the relight branch using adversarial loss with a small learning rate $1e-6$. The parameters of discriminator are fixed.
\end{itemize}

\section{More Experiments} 
\noindent\textbf{Comparison with FaceLit}
We also provide comparisons with FaceLit~\cite{ranjan2023facelit} in Fig~\ref{fig:facelit}. FaceLit is unsupervised disentanglement method based on volume rendering, which struggle with extreme lighting conditions and often produce unrealistic lighting artifacts. Moreover, it relies on a white-light assumption and therefore cannot handle colored lighting. In contrast, our method supports diverse lighting inputs, including color environment maps, point lights, and spherical harmonic coefficients. It also produces natural lighting and shadow effects even under strong side lighting conditions.

\begin{table}[t!]
    \centering
    \begin{tabular}{c|c}
    \toprule
    Finetune & Identity Similarity\\
    \midrule
        Layer 6 &  0.9685 \\
        Layer 5+6 & 0.9654\\
        All Layers & 0.9599 \\
    \bottomrule
    \end{tabular}
    \caption{Quantitative comparison of identity preservation when fine-tuning different StyleGAN layers, measured by average cosine similarity (higher is better) between generated faces and the original GGHead output using the face\_recognition library.}
    \label{tab:id_sim}
\end{table}

\noindent\textbf{Ablation Study on Base Branch Adaption.}
We quantitatively evaluated identity preservation when fine-tuning different layers of StyleGAN. Specifically, we randomly generated 100 images and used the face\_recognition library to compute the facial embedding for each image. We then measured the similarity between images produced by GGHead and those generated by three variants using the same latent codes: (1) fine-tuning only layer 6; (2) fine-tuning layers 5 and 6; and (3) fine-tuning all layers. As shown in the table~\ref{tab:id_sim}, the results indicate that fine-tuning the entire model significantly alters facial identity, whereas fine-tuning only layer 6 or layers 5+6 preserves identity to a similar degree. However, qualitative results show that optimizing only layer 6 still introduces noticeable lighting artifacts. Therefore, we chose to optimize layers 5 and 6 jointly.

\section{Limitations.} The model is static and does not support the control of facial expressions. Furthermore, the number of captured identities and lighting conditions remains limited, which restricts the model's ability to learn a more comprehensive prior over diverse facial material properties. Future work could address these by incorporating video captures and expanding the dataset scale.

\section{Ethical Considerations}
The generation of artificial portrait using our method poses risks, including the spread of false information, and erosion of trust in media credibility. These issues could have profound societal implications. Addressing this challenge requires developing reliable techniques to identify and verify authentic content.

\clearpage

\end{document}